\pdfoutput=1
\documentclass{article}

\usepackage{microtype}
\usepackage{graphicx}
\usepackage{booktabs} 
\usepackage[T1]{fontenc}

\usepackage{hyperref}

\usepackage{amssymb}
\usepackage{microtype}
\usepackage{graphicx}
\usepackage{subcaption}

\usepackage{amsfonts}
\usepackage{amsthm}
\usepackage{amsmath}
\usepackage{soul}
\usepackage{algorithm2e}

\newcommand{\fig}[1]{Figure~\ref{fig:#1}}
\newcommand{\sect}[1]{Section~\ref{sect:#1}}
\newcommand{\tab}[1]{Table~\ref{tab:#1}}
\newcommand{\alg}[1]{Algorithm~\ref{alg:#1}}
\newcommand{\eq}[1]{(\ref{eq:#1})}

\usepackage{enumitem}

\usepackage[accepted]{icml2019}

\icmltitlerunning{Learning to Route in Similarity Graphs}

\begin{document}

\twocolumn[
\icmltitle{Learning to Route in Similarity Graphs}



\icmlsetsymbol{equal}{*}

\begin{icmlauthorlist}
\icmlauthor{Dmitry Baranchuk}{yandex,msu}
\icmlauthor{Dmitry Persiyanov}{mipt}
\icmlauthor{Anton Sinitsin}{yandex,hse}
\icmlauthor{Artem Babenko}{yandex,hse}
\end{icmlauthorlist}

\icmlaffiliation{yandex}{Yandex, Russia}
\icmlaffiliation{mipt}{Moscow Institute of Physics and Technology, Russia}
\icmlaffiliation{msu}{Lomonosov Moscow State University, Russia}
\icmlaffiliation{hse}{National Research University Higher School of Economics, Russia}

\icmlcorrespondingauthor{Dmitry Baranchuk}{dmitry.baranchuk@graphics.cs.msu.ru}
\icmlkeywords{Machine Learning, ICML}

\vskip 0.3in
]



\printAffiliationsAndNotice{}  

\begin{abstract}
Recently similarity graphs became the leading paradigm for efficient nearest neighbor search, outperforming traditional tree-based and LSH-based methods. Similarity graphs perform the search via greedy routing: a query traverses the graph and in each vertex moves to the adjacent vertex that is the closest to this query. In practice, similarity graphs are often susceptible to local minima, when queries do not reach its nearest neighbors, getting stuck in suboptimal vertices. In this paper we propose to learn the routing function that overcomes local minima via incorporating information about the graph global structure. In particular, we augment the vertices of a given graph with additional representations that are learned to provide the optimal routing from the start vertex to the query nearest neighbor. By thorough experiments, we demonstrate that the proposed learnable routing successfully diminishes the local minima problem and significantly improves the overall search performance.
\end{abstract}

\section{Introduction}
\label{sect:intro}
Nearest neighbor search (NNS) is an extensively used subroutine in a whole range of machine learning systems for non-parametric classification/regression, language modeling, information retrieval, recommendations and others. 
Modern applications have to work with vast data volumes, hence the scalability of the NNS approaches became a problem of great interest for the machine learning community. Formally the NNS problem is stated as follows. Given the database $S = \{v_1,\dots,v_N\} \subset \mathbf{R}^D$ and a query $q \in \mathbf{R}^D$, one needs to find the datapoint $v \in S$ that is the closest to the query in terms of some metric (e.g. Euclidean distance). 

The current approaches for efficient NNS mostly belong to three separate lines of research. The first family of methods, based on partition trees \cite{KdTree,PcaTree,ApdTree,RpTree,dasgupta2013randomized}, hierarchically split the search space into a large number of regions, corresponding to tree leaves, and query visits only a limited number of promising regions when searching. The second, locality-sensitive hashing methods \cite{LSH98, LSH, andoni2008near, Razenshteyn15} map the database points into a number of buckets using several hash functions such that the probability of collision is much higher for nearby points than for points that are further apart. At the search stage, a query is also hashed, and distances to all the points from the corresponding buckets are evaluated. Finally, similarity graphs methods \cite{navarro2002searching,malkov2018efficient,fu2016efanna,NSG} represent the database as a graph, and on the search stage, a query traverses the graph via greedy exploration. The empirical performance of similarity graphs was shown to be much higher compared to LSH-based and tree-based methods\cite{malkov2018efficient}, and our paper falls in this line of work on NNS.

In more details, the typical search process in similarity graphs performs as follows. The database is organized in a graph, where each vertex corresponds to some datapoint, and the vertices, corresponding to the neighboring datapoints, are connected by edges. The search algorithm picks a vertex (random or predefined) and iteratively explores the graph from there. On each iteration, the query tries to greedily improve its position via moving to an adjacent vertex that is closest to the query. The routing process stops when there are no closer adjacent vertices, or the runtime budget is exceeded.

It was shown\cite{navarro2002searching} that if the similarity graph contains all the edges from the Delaunay graph, constructed for the database $S$, then the greedy routing, described above, is guaranteed to find the exact nearest neighbor. However, for high-dimensional data, both storage and traversal of the full Delaunay graph would be infeasible, due to a very high number of edges\cite{Beaumont07}. Hence, the state-of-the-art practical methods use approximate Delaunay graphs, restricting maximal vertex degrees by a fixed value. Unfortunately, this approximation often results in the problem of local minima, when the graph traversal gets stuck in a suboptimal vertex, which has no neighbors, that are closer to query.

In this paper we claim that the local minima problem is caused mainly by the fact that the routing decisions are made locally in each vertex, and do not explicitly account the graph global structure. Our approach aims to overcome this issue by learning the routing function for a given similarity graph. In more details, we augment each vertex with an additional compact representation that is used for the routing decision on the search stage. These representations are learned via explicit maximization of optimal routing probability in a given similarity graph, hence explicitly consider both query distribution and the global graph structure. Furthermore, we observe that the dimensionality of these representations could often be smaller than the original data dimensionality, which improves the routing computational efficiency.

Overall, we summarize the contributions of this paper as follows:
\begin{enumerate}
    \item We propose an algorithm to learn the routing function in the state-of-the-art similarity graphs. The algorithm explicitly accounts the global graph structure and reduces the problem of local minima.
    \item We experimentally demonstrate that the proposed learnable routing substantially increases the search accuracy on three open-source datasets for the same runtime budget.
    \item The PyTorch source code of our algorithm is available online\footnote[1]{https://github.com/dbaranchuk/learning-to-route}.
\end{enumerate}

The rest of the paper is organized as follows. We discuss related work in section \ref{sect:related} and present the proposed learnable routing in section \ref{sect:theory}. We present our experimental evaluations in section \ref{sect:experiments} and conclude in section \ref{sect:conclusion}.

\section{Related work}
\label{sect:related}
In this section we review the main ideas from the existing works that are relevant to our approach and will be used in the description of our method.

\textbf{Nearest neighbor search problem.} The problem of nearest neighbor search is well-known for the machine learning community for decades. Two established lines of research on the NNS problem include LSH and partition trees methods. These families of methods have strong theoretical foundations and allow to estimate the search time or the probability of successful search\cite{andoni2008near,dasgupta2013randomized}. Recently, the paradigm of similarity graphs proved itself to be efficient for NNS. While similarity graphs do not provide solid theoretical guarantees yet, their empirical performance appears to be much higher compared to trees or LSH\cite{malkov2018efficient}.

\textbf{Similarity graphs.} For a database $S = \{v_i \in \mathbf{R}^D | i = 1,\dots,n\}$ the similarity graph is a graph, where each vertex corresponds to one of the datapoints $v$. The vertices $v_i$ and $v_j$ are connected by an edge if $v_j$ belongs to the set of $k$ nearest neighbors of $v_i$, $v_j \in NN_k(v_i)$ in terms of some metric. The search in such a graph is performed via \textit{greedy routing}. A query starts from a random vertex and then on each step moves from the current vertex to its neighbor that appears to be the closest to a query. The process terminates when the query reaches a local minimum or the runtime budget is exceeded.

The process, described above, was initially proposed in the seminal paper\cite{navarro2002searching} that gave rise to research on NNS with similarity graphs. Since then plenty of methods, which elaborate the idea, were developed\cite{malkov2018efficient, fu2016efanna, NSG}. In this paper we aim to improve the routing in one of the recent graphs, Hierarchical Navigable Small World (HNSW)\cite{malkov2018efficient}, as it is shown to provide the state-of-the-art performance on the common benchmarks and its code is publicly available. Note that other types of similarity graphs could use the proposed learnable routing as well.

When searching, the HNSW similarity graph\cite{malkov2018efficient} maintains a priority queue of size $L$ with the graph vertices, which neighbors should be visited by the search process. With $L{=}1$ the search is equivalent to greedy routing, while with $L>1$ it can be considered as Beam Search\cite{Bisiani87}, which makes the search process less greedy. Typically varying $L$ determines the trade-off between the runtime and search accuracy in similarity graphs.

\textbf{Learning to search.} Learning to search\cite{Daume} is a family of methods for solving structured prediction problems by learning to navigate the space of possible solutions. They operate by introducing a parametric model of the search procedure and tuning its parameters to fit the optimal search strategy. 

These methods have seen numerous applications to tasks such as Part Of Speech Tagging\cite{Daume_better}, Machine Translation\cite{bso_good,bso_2018_bad}, Scene Labelling\cite{scene_labelling} and others.

Learning to search can be viewed as an extension of Imitation Learning\cite{dagger,gail} methods: a search "agent" is trained to follow the expert search procedure. The expert is an algorithm that optimally solves the specific search problem, e.g. produces the best possible translation. While not always accessible, such expert decisions can usually be computed for labeled training data points using any exact search algorithm.

To the best of our knowledge, Learning to Search has not yet been applied to the task of approximate nearest neighbor search. However, it appears to be a natural fit for searching in similarity graphs discussed above.

\section{Method}
\label{sect:theory}
We propose a model that directly learns to find nearest neighbors with a given similarity graph. To do so, we reformulate the graph routing algorithm as a probabilistic model and train it by maximizing the probability of optimal routing for a large set of training queries. 

\subsection{Stochastic Search Model}

The typical way to navigate in a similarity graph is through beam search (\alg{beam_search}). In its essence, it is an algorithm that iteratively expands the nearest vertex from a heap of visited vertices. The process stops when the heap becomes empty, or the runtime budget is exceeded. In this paper we focus on the latter budgeted setting with the limit of distance computations, specified by the user. 
\begin{algorithm}[H]
 \caption{Beam search}
 \KwData{graph G, query $q$, initial vertex $v_0$, output size $k$}
 \textbf{Initialization:}
 
 V = $\{v_0\}$  // a set of visited vertices
 
 H = $\{v_0 : d(v_0, q)\}$  // a heap of candidates
                            
 \While{has runtime budget}{
    $v_i$ = SelectNearest(H, q)
    
    \For{$\hat v \in Expand(v_i, G)$}{
        \If{$\hat v \not\in V$}{
            $V := Add(V, \hat v)$
            
            $H := Insert(H, \hat v, d(\hat v, q))$
        }
    }
}
\Return TopK(V, q, k)
\label{alg:beam_search}
\end{algorithm}
We generalize this algorithm into a stochastic search: instead of selecting a vertex that has the smallest distance to a query, stochastic search samples the next vertex from a softmax probability distribution over vertices in the current heap $H$:
\vspace{-1mm}
\begin{equation} \label{eq:probability_of_vertex}
P(v_i | q, H) = {e ^ {- d(v_i, q) / \tau} \over \sum\limits_{v_j \in H} e ^ {- d(v_j, q) / \tau}}
\end{equation}
\vspace{-2mm}

Once stochastic search terminates, it samples $k$ visited vertices from the softmax distribution over the set of visited vertices $V$ instead of $H$. Those vertices are returned as the search result. If $\tau \rightarrow 0^{+}$, one recovers the original beam search \alg{beam_search}. Under $\tau > 0$, the algorithm may sometimes pick suboptimal vertices.

Our core idea is to replace the distance $d(v_i, q)$ in the original data space with a negative inner product between learnable mappings $-\langle f_\theta(v_i), g_\theta(q) \rangle$, resulting in:
\begin{equation} \label{eq:probability_parametric}
P(v_i | q, H, \theta) = {e ^ {\langle f_\theta(v_i), g_\theta(q) \rangle} \over \sum\limits_{v_j \in H} e ^ {\langle f_\theta(v_j), g_\theta(q) \rangle}}
\end{equation}
\vspace{-2mm}

Here, $f_\theta(\cdot)$ is a neural network for database points and $g_\theta(\cdot)$ is another neural network for queries. The network parameters $f_\theta$ and $g_\theta$ are jointly trained in a way that helps stochastic search to reach the actual nearest neighbor $v^*$ and to return it as one of $k$ output datapoints. We discuss the actual network architectures and the optimization details below. Note that the \alg{beam_search} with our learnable routing returns top-$k$ based on the values of inner products $\langle f_\theta(v_i), g_\theta(q) \rangle$ while the original NNS problem requires to find the nearest Euclidean neighbors. Therefore, as a final search step, we additionally rerank the retrieved top-$k$ datapoints based on the Euclidean distances to the query in the original space.

\subsection{Optimal routing}
In order to train our model to pick the optimal vertices, we introduce the notion of \textit{optimal routing} --- a routing that follows the shortest path from the start vertex to the actual nearest neighbor $v^*$ and returns it among top-$k$ candidates for reranking. For simplicity, we assume that the computational budget is large enough for the algorithm to reach $v^*$ under such routing.

For a formal definition of optimal routing, consider an oracle function $Ref(H)$. This function selects vertices from $H$ that are the closest to the actual nearest neighbor $v^*$ in terms of hops over the graph edges. A sequence of vertices is an optimal routing iff it expands $v_i \sim Ref(H)$ on each iteration until it finds the actual nearest neighbor. Once $v^*$ is found, the algorithm should select $Ref(V)$ as one of top-$k$ vertices. 

In practice, the values of $Ref(H)$ for the training queries are obtained as follows. We compute the distances (in terms of graph hops) to $v^*$ from each vertex in $H$ via simple Breadth-First Search (BFS) algorithm and then return the vertex, corresponding to the minimal number of hops. In order to improve the training performance, we precompute the hop distances for each training query before the training begins. We store the pre-computed distances for all training queries in a persistent cache and access them on the fly as the training progresses. In order to optimize memory requirements, we only store distances to $v^*$ from the vertices that are likely to be visited by a search with the corresponding query. We select those vertices with a simple heuristic: a vertex is selected if there exists a near-optimal path from the start vertex to the actual neighbor $v^*$ that goes through that vertex. In particular, we consider all vertices along the paths that are at most $m=5$ hops longer than the optimal path from the start vertex to $v^*$.

\subsection{Training objective}

We train our probabilistic search model\eq{probability_parametric} to perform the optimal routing. The naive approach would be to explicitly maximize the log-likelihood of optimal routing:

\begin{flalign}\label{eq:mle}
\begin{split}
    & J_{naive} = \underset{q, v^*}E \log P( Opt(q) | q, \theta) = \\
    & \underset{q, v^*}E \sum_{\substack{v_i, H_i \in \\ Opt(q)}}
    log P(v_i | q, H_i, \theta)
    + log P(v^* \in TopK | q, V, \theta))
\end{split}
\vspace{-0.2cm}
\end{flalign}
where $v_i, H_i \in Opt(q)$ stands for iterating over vertices and heap states on each step of optimal routing for $q$ and $ P(v^* \in TopK | q, V, \theta) $ is a probability of $v^*$ being selected as one of the top-$k$ datapoints that the routing algorithm visits. If the actual nearest neighbor $v^*$ belongs to the top-$k$ when the overall search for this query will be successful as the top-$k$ datapoints are reranked based on the original distances.

Maximizing the objective \eqref{eq:mle} would only train the algorithm to search on the vertices from the optimal routing trajectories. However, when applied on unseen queries, the routing learned in this way could result in poor performance. Once this search algorithm makes an error (i.e., picks a non-optimal vertex) at any routing step, it adds the wrong vertex to its heap $H$. Therefore, after a single error, the search algorithm enters a state that never occurred during training. In other terms, the algorithm is not trained to compensate for its errors, and a single mistake will likely ruin all future subsequent routing steps.

In order to mitigate this issue, we change the training procedure, making it close to the paradigm of Imitation Learning\cite{imit}. Intuitively, we allow the algorithm to route the graph based on the current parameters of $f_\theta(\cdot)$ and $g_\theta(\cdot)$ and possibly choose suboptimal vertices. When search stops, we update the parameters $\theta$ to force the algorithm to follow the optimal routing in each visited vertex despite previous mistakes.

Formally, we maximize the following objective:
\vspace{-0.2cm}
\begin{flalign}\label{eq:imitation_objective}
\begin{split}
J_{imit} = &\underset{q, v^*}E \underset{\substack{v_i, H_i \in \\ Search_\theta(q)}}\sum log P(v_i \in Ref(H_i) | q, H_i, \theta) + \\ 
   & + log P(v^* \in TopK | q, V, \theta))
\end{split}
\end{flalign}
\vspace{-0.2cm}

Here, $v_i, H_i \in Search_\theta(q)$ denotes the sequence of vertices and corresponding heap states that occur during search for a query $q$ with the routing defined by $f_\theta(\cdot)$ and $g_\theta(\cdot)$ with parameters $\theta$. Note that this is different from \eqref{eq:mle} where we would only consider trajectories under the optimal routing.

When maximizing \eq{imitation_objective}, the gradients of the first terms $log P(v_i \in Ref(H_i) | q, H_i, \theta)$ are obtained via differentiating \eq{probability_parametric} w.r.t. $\theta$. The second term is a probability that ground truth vertex $v^*$ is chosen among top-$k$ nearest candidates after search terminates. Formally, this is a probability that $v^*$ will be chosen among k candidates sampled from \eq{probability_parametric} without replacement. Unfortunately, computing this probability exactly requires iterating over all possible selections of top-k elements, which is intractable for large $k$. Instead, we use an approximation, similar to that of \cite{wiseman2016sequence}. Namely, we sample $k - 1$ vertices from $v_0, ..., v_{k-1} \sim P(v_i | q, V \setminus \{v^*\})$ without replacement and compute the probability that $v^*$ will be sampled from what's left: $P(v^* | q, V \setminus \{v_0, ..., v_{k-1}\} )$.

Our approach can be considered as a special case of DAGGER algorithm\cite{dagger} for imitation learning. Indeed, in terms of imitation learning, the stochastic search algorithm \eq{probability_parametric} defines an agent. This agent is trained to imitate the "expert" decisions that are the optimal routes, which are precomputed by BFS.

\vspace{-0.3cm}
\begin{figure}[H]
    \centering
    \includegraphics[width=200px,height=100px]{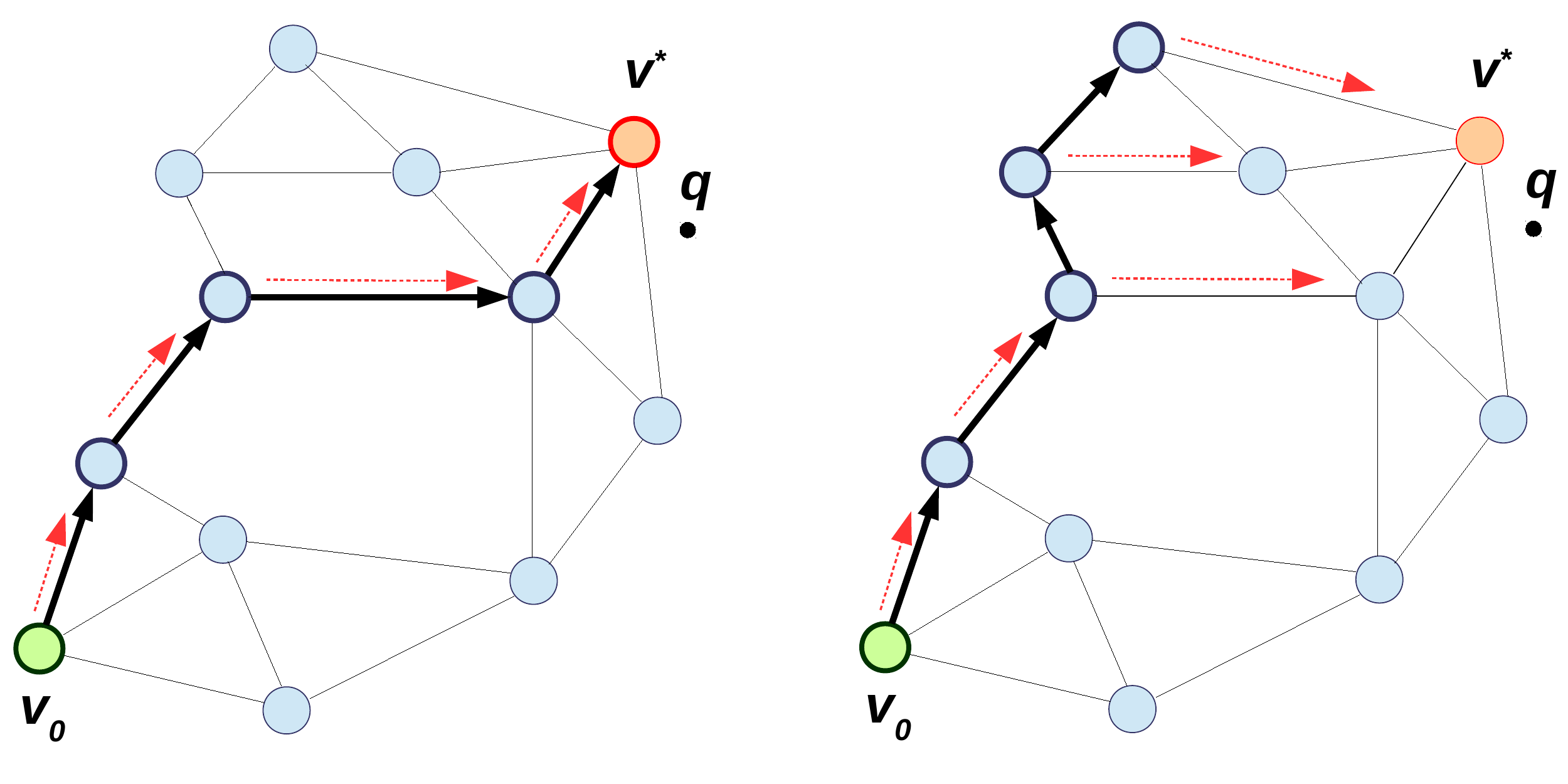}
    \vspace{-4mm}
    \caption{\textbf{Left:} teacher forcing, objective is added up over optimal routing. The sequence of visited vertices is drawn in bold. Training supervision is shown with red arrows. \textbf{Right:} imitation learning, model made an error at step 3 and diverged from optimal routing. Objective is computed from the vertices visited by the model.}
    \label{fig:teacher_forcing}
\end{figure}
\vspace{-3mm}
The difference between the two objectives \eq{mle} and \eq{imitation_objective} is visually demonstrated on \fig{teacher_forcing}. The objective \eq{mle} is akin to the Teacher Forcing\cite{williams1989learning}, when only the vertices from the optimal route participate in the objective. In contrast, \eq{imitation_objective} corresponds to Imitation Learning\cite{imit} setup, when the routing is performed with the current parameter values, then in each of the visited vertices "expert" tells how the parameters should be changed to get to the optimal route. 

\subsection{Model Architecture}\label{sect:model}

\begin{figure}
\noindent
\centering
\renewcommand\arraystretch{0.8}
\begin{tabular}{c}
\includegraphics[height=2cm]{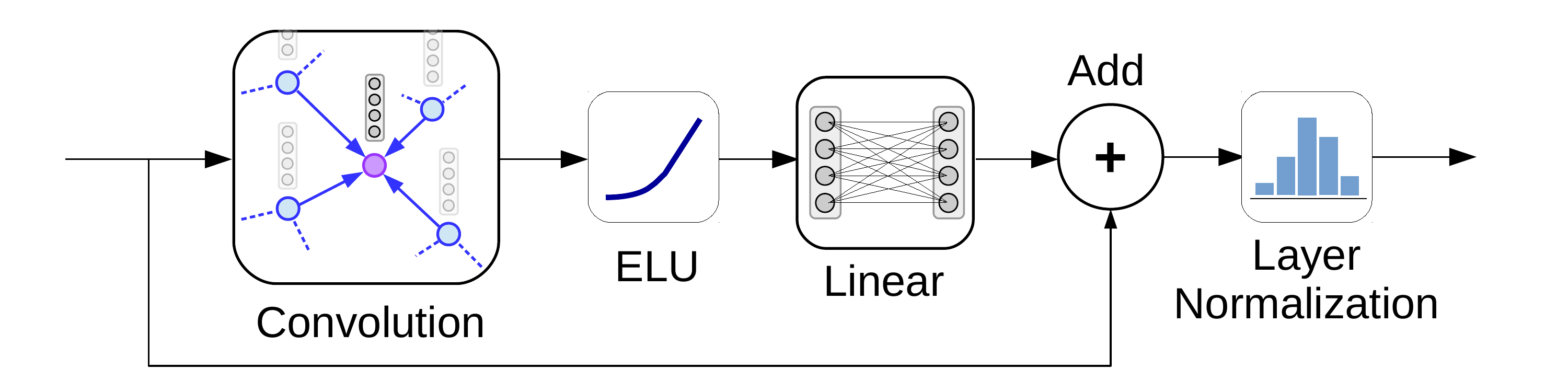}\\
\includegraphics[height=2cm]{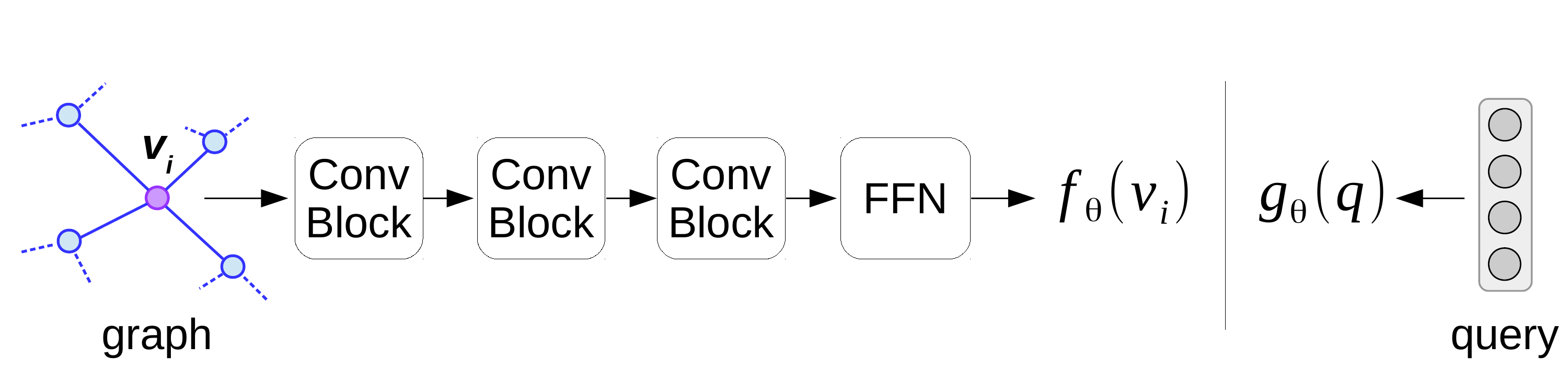}
\end{tabular}
\vspace{-3mm}
\caption{The network architecture used in the most of our experiments. \textbf{Top:} the Conv block, which consists of graph convolution layer, the ELU nonlinearity and the fully-connected layer. The residual connection goes through it. The block ends with the layer normalization. \textbf{Bottom-left:} the database branch $f_\theta(\cdot)$, which includes three Conv blocks followed by a feed-forward network (FFN) consisting of two fully-connected layers with ELU nonlinearity. \textbf{Bottom-right:} the query branch $g_\theta(\cdot)$, which is usually an identity transformation or a linear mapping.}
\label{fig:arch}
\end{figure}

The algorithm described above allows for arbitrary differentiable functions for $f$ and $g$. This opens a wide range of possibilities, e.g. linear projections, feed-forward neural networks, Graph Neural Networks\cite{GNN}. The architecture we used in our experiments is presented on \fig{arch}.

Our architecture is asymmetric, i.e. $f_\theta(\cdot)$ and $g_\theta(\cdot)$ are different and do not share parameters. The database branch $f_\theta(\cdot)$ contains three Graph Convolutional layers\cite{gcn} with ELU nonlinearity\cite{elu}, as well as Layer Normalization\cite{ln} and residual connections\cite{he2016deep} for faster convergence. Note that $f_\theta(\cdot)$ can be of any computational complexity as the vertex representations $f_\theta(v)$ are precomputed offline. In contrast, the query branch $g_\theta(\cdot)$ should be computationally efficient as it is computed online for a query before the search process starts. In this paper we experiment with two options for $g_\theta(\cdot)$:
\begin{itemize}[leftmargin=2mm,itemindent=.5cm,labelwidth=\itemindent,labelsep=0cm,align=left]
\itemsep-0.1em
    \item $g_\theta(q) = q$, identity transformation. In this case $g_\theta(\cdot)$ does not require additional computational cost.
    \item $g_\theta(q) = W\times q$, where $W \in \mathbf{R}^{d\times D}$. In this case, both 
    $f_\theta(v), g_\theta(q) \in \mathbf{R}^d$. If $d < D$, the routing becomes more efficient, as the computation of inner products $\langle f_\theta(\cdot), g_\theta(q) \rangle$ requires $O(d)$ operations. On the other hand, this option requires $O(d\times D)$ preprocessing for the queries on the search stage.
\end{itemize}

Our stochastic search model \eq{probability_parametric} is trained using minibatch gradient descent algorithm on the routing trajectories sampled from the probability distribution it induces with the current parameters $\theta$. In all the experiments we use Adam\cite{kingma2014adam} algorithm for SGD training. We have also observed significantly faster convergence from One Cycle learning rate schedule\cite{smith2017super}.





\subsection{Search}

Once the model is trained, we precompute $f_\theta(v_i)$ for all the database points offline, while the queries are transformed $q \rightarrow g_\theta(q)$ on-the-fly. The search process is performed in the same way as \alg{beam_search}, the only difference being that the routing decisions are based on the inner products $\langle f_\theta(\cdot), g_\theta(q) \rangle$ instead of the original Euclidean distances. After routing stops, top-$k$ visited vertices, corresponding to the largest values of $\langle f_\theta(\cdot), g_\theta(q) \rangle$ are reranked based on the Euclidean distances to the query in the original space.

\subsection{Scalability}

As will be shown below, most of our experiments were performed on graphs with 100,000 vertices and 100,000 training queries. On this scale pre-computation of $Ref(v)$ functions for the training queries takes approximately 20 minutes on a machine with 12 CPU cores. The DNN training on a single 1080Ti GPU takes on average twelve hours for the architecture described in \ref{sect:model}. We expect that the training of our algorithm can be scaled to larger graphs with an estimated linear increase in training time. One could also boost the training performance with multi-gpu or using the techniques for Graph Neural Network acceleration\cite{fastgcn}. However, we did not perform such experiments within this study.

\section{Experiments}
\label{sect:experiments}
\subsection{Toy example}
\begin{figure}[ht]
\vskip 0.2in
\begin{center}
\centerline{\includegraphics[width=\columnwidth]{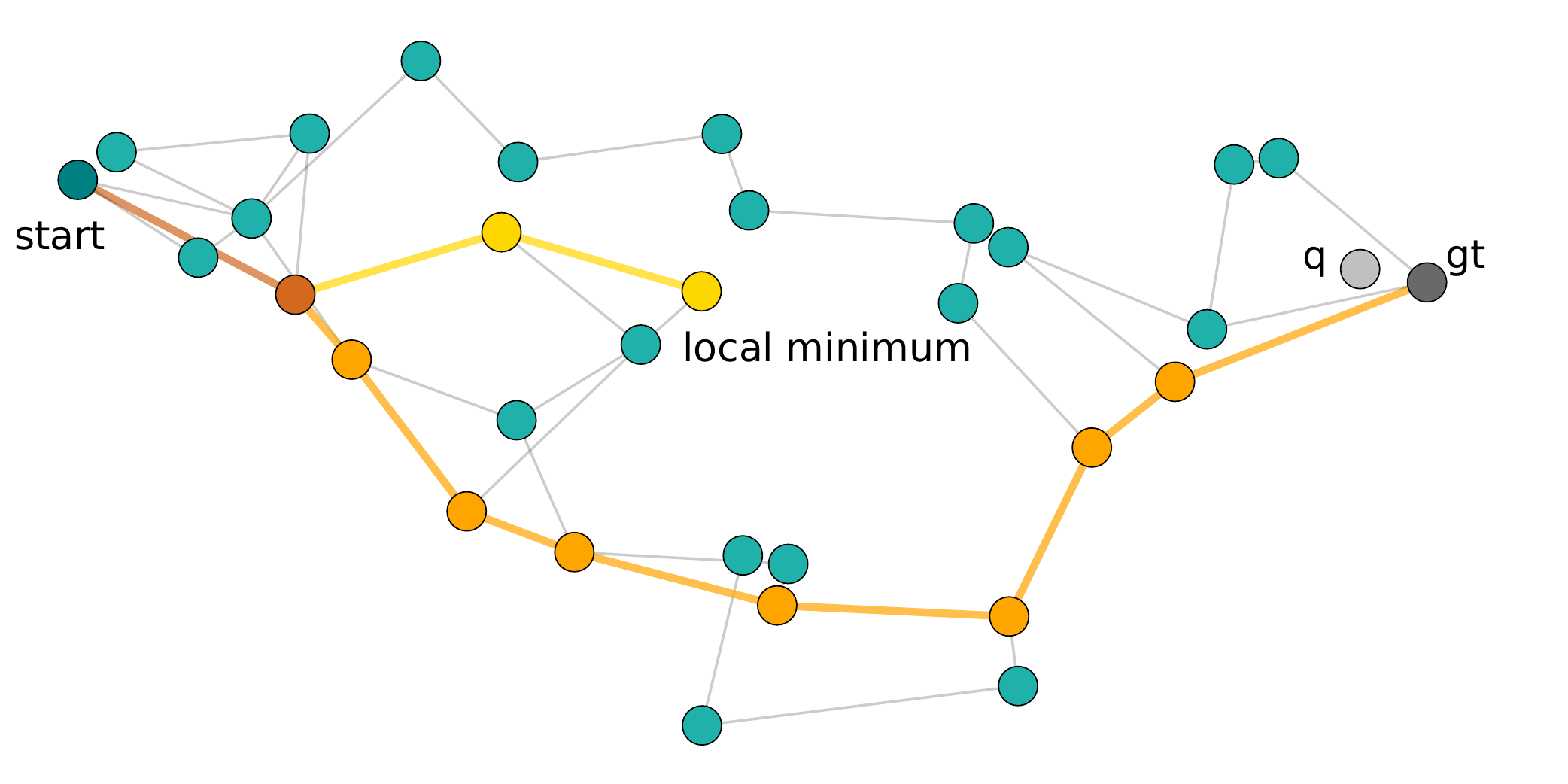}}
\vspace{-5mm}
\caption{The examples of the greedy routing based on the original data and the learned representations. The database of 33 datapoints is organized in a similarity graph. When searching with a query $q$ (shown in grey), the greedy routing based on the original distances (shown in yellow) does not find the groundtruth (shown in dark grey) falling in a local minimum. On the contrary, the greedy routing based on the learned representations (shown in orange) decently reaches the nearest neighbor.}
\label{fig:toy}
\end{center}
\vskip -0.2in
\end{figure}

We start by a toy experiment to demonstrate the problem of local minima and the advantage of the proposed learnable routing. In this experiment we have a small database of 33 two-dimensional points, organized in a similarity graph as shown on \fig{toy}. For a query $q$, the greedy routing based on the original datapoints (the yellow edges) gets stuck in a local minimum and does not reach the actual nearest neighbor. In contrast, the greedy routing based on the learned representations (the orange edges) successfully visits the groundtruth. In this toy experiment we take $f_\theta(\cdot)$ being a simple two-layer perceptron with the hidden layer size $128$.   

\subsection{Problem and datasets}

In this paper we focus on the budgeted nearest neighbor search\cite{yu2017greedy}, i.e. when the user specifies the limit on the number of computations. In particular, we set the maximal number of distance computations \textit{(DCS)} and compare the performance of different methods under this budget. As the primary performance measure, we use the $Recall@R$, which is calculated as a rate of queries for which the true nearest neighbor is presented within the top $R$ candidates.

In the experiments below we set $DCS=128,256,512$ to investigate the routing performance in low, medium and high $Recall@1$ niches respectively. Note that the proposed learnable routing requires a separate training for each particular $DCS$ value, that allows the vertex representations to adapt to the particular problem setup. We always learn the routing representations on top of the bottom layer of the Hierarchical Navigable Small World graph\cite{malkov2018efficient}, which we refer to as NSW.



We evaluate the proposed approach on three publicly available datasets, which are widely used as the large-scale nearest neighbor search benchmarks:
\vspace{-3mm}

\begin{enumerate}[leftmargin=2mm,itemindent=.5cm,labelwidth=\itemindent,labelsep=0cm,align=left]
\itemsep-0.1em
    \item SIFT100K dataset\cite{Jegou11a} is sampled from one million of $128$-dimensional SIFT descriptors. We consider 100,000 learn vectors as train queries. The hold-out 10,000 query vectors are used for evaluation.
    \item DEEP100K dataset\cite{BabenkoCVPR16} is a random 100,000 subset of one billion of $96$-dimensional CNN-produced feature vectors of the natural images from the Web. We sample 100,000 train queries from the learn set. For evaluation we take original 10,000 queries.
    \item GloVe100K dataset\cite{pennington2014glove} is a collection of $300$-dimensional normalized $GloVe$ vector representations for \textit{Wikipedia 2014 + Gigaword 5}. We randomly split the original 400,000 word embeddings on base and learn sets, each containing 100,000 vectors. 10,000 queries are taken from the remaining vectors for evaluation. 
\end{enumerate}
\vspace{-3mm}
For each dataset we construct the NSW graph on the base set with the optimal maximal vertex out-degree $MaxM{=}16$ and learn the routing representations for $DCS=128,256,512$ as described in \sect{theory}.

\subsection{Routing evaluation}
\begin{figure*}
\noindent
\centering
\begin{tabular}{ccc}
\vspace{-1mm}
\includegraphics[height=4.4cm]{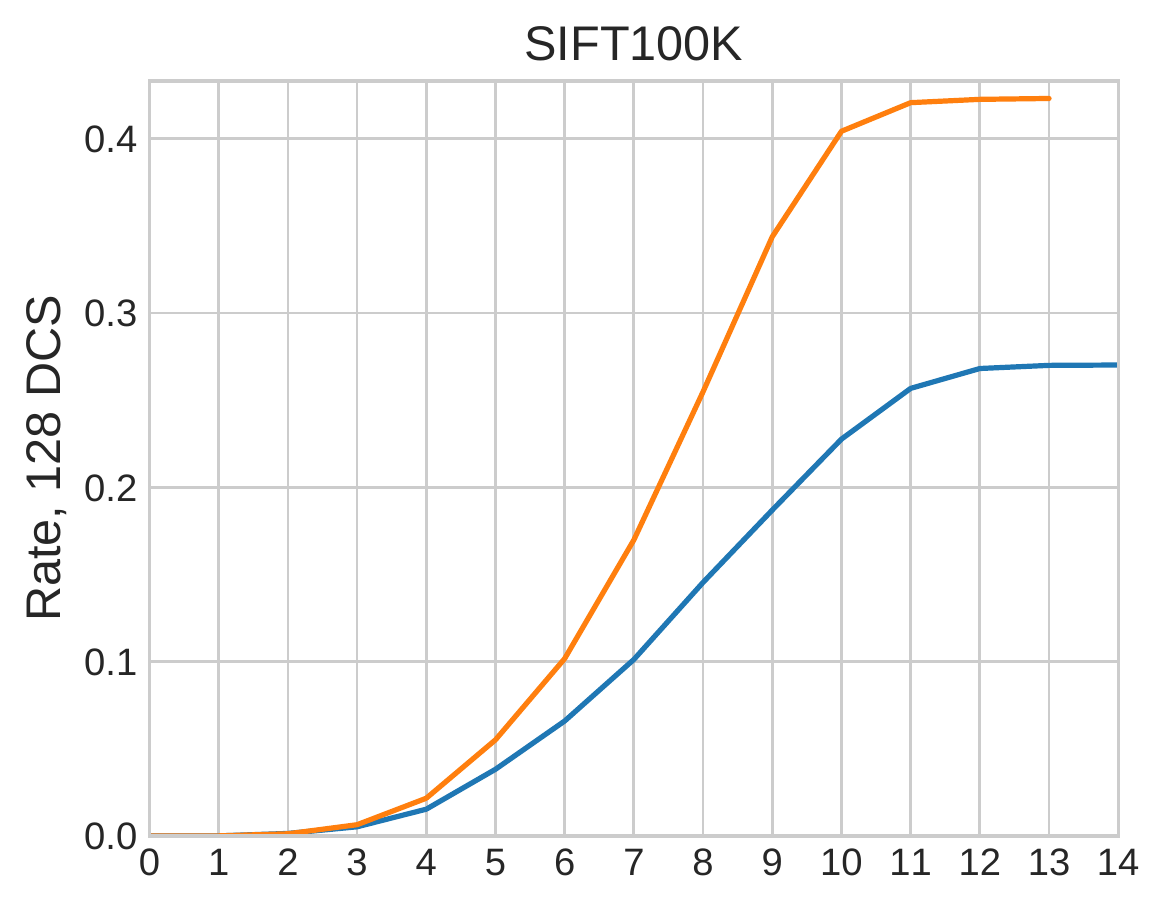}&
\includegraphics[height=4.4cm]{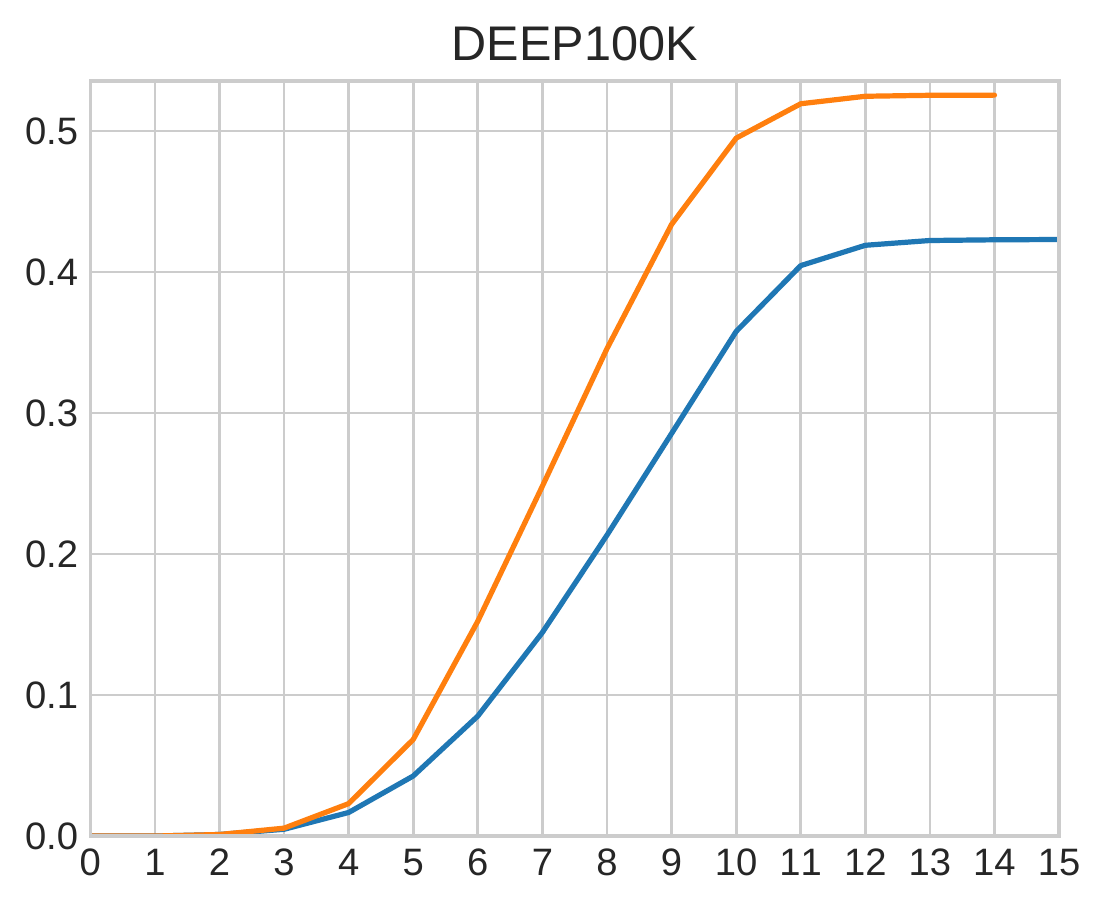}&
\includegraphics[height=4.4cm]{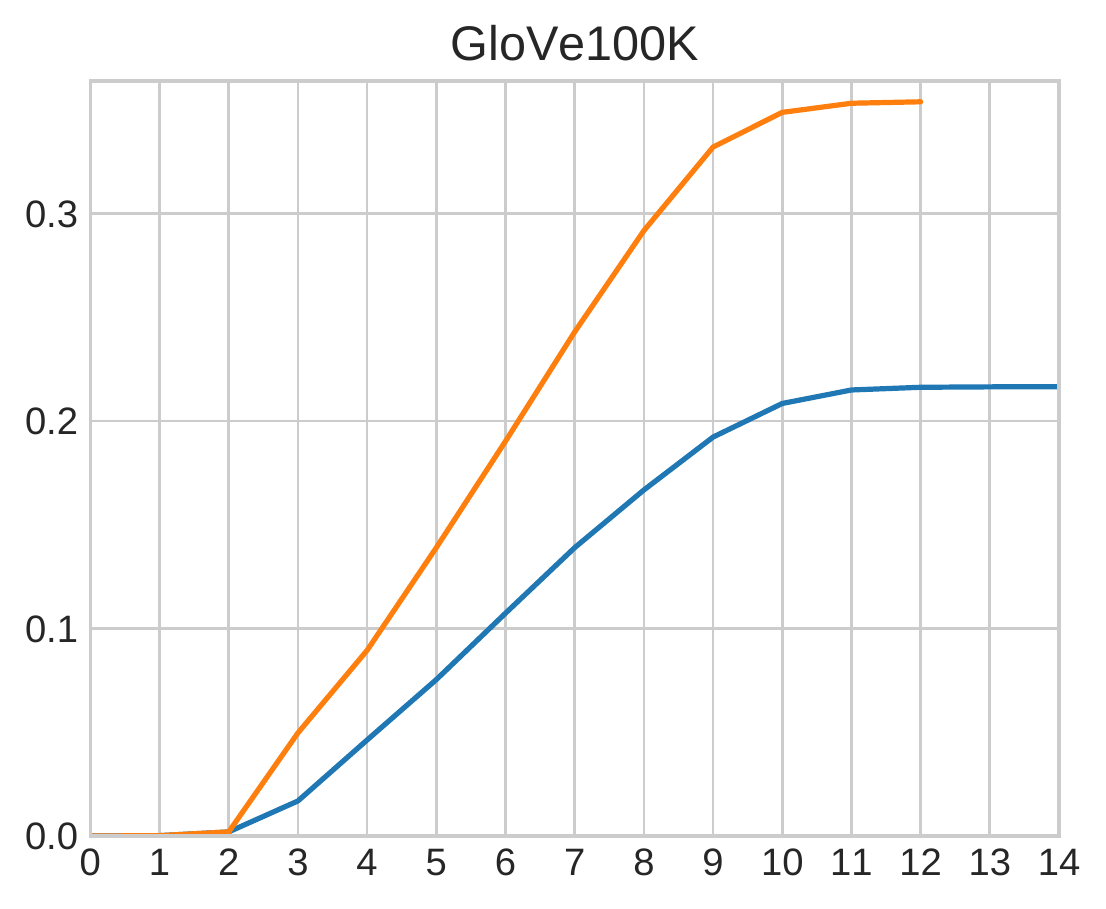} \\
\includegraphics[height=4.4cm]{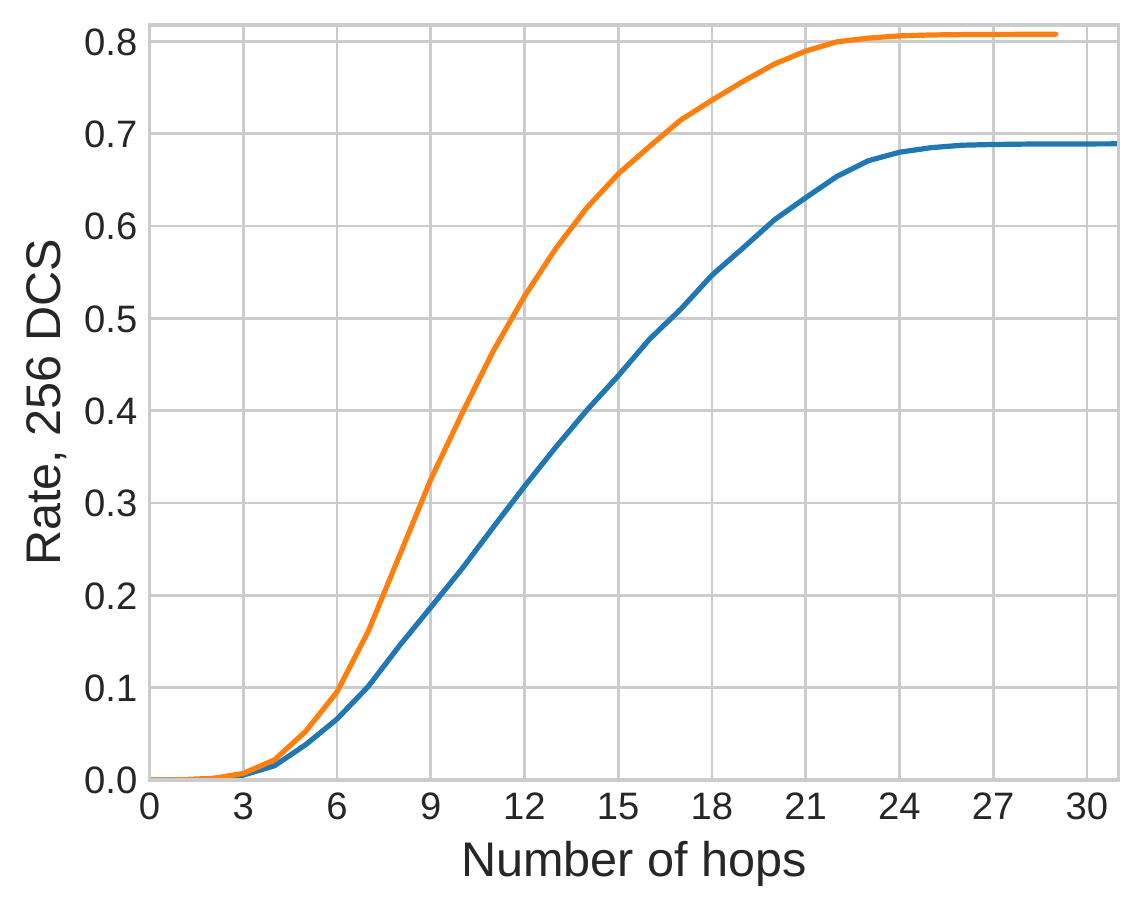}&
\includegraphics[height=4.4cm]{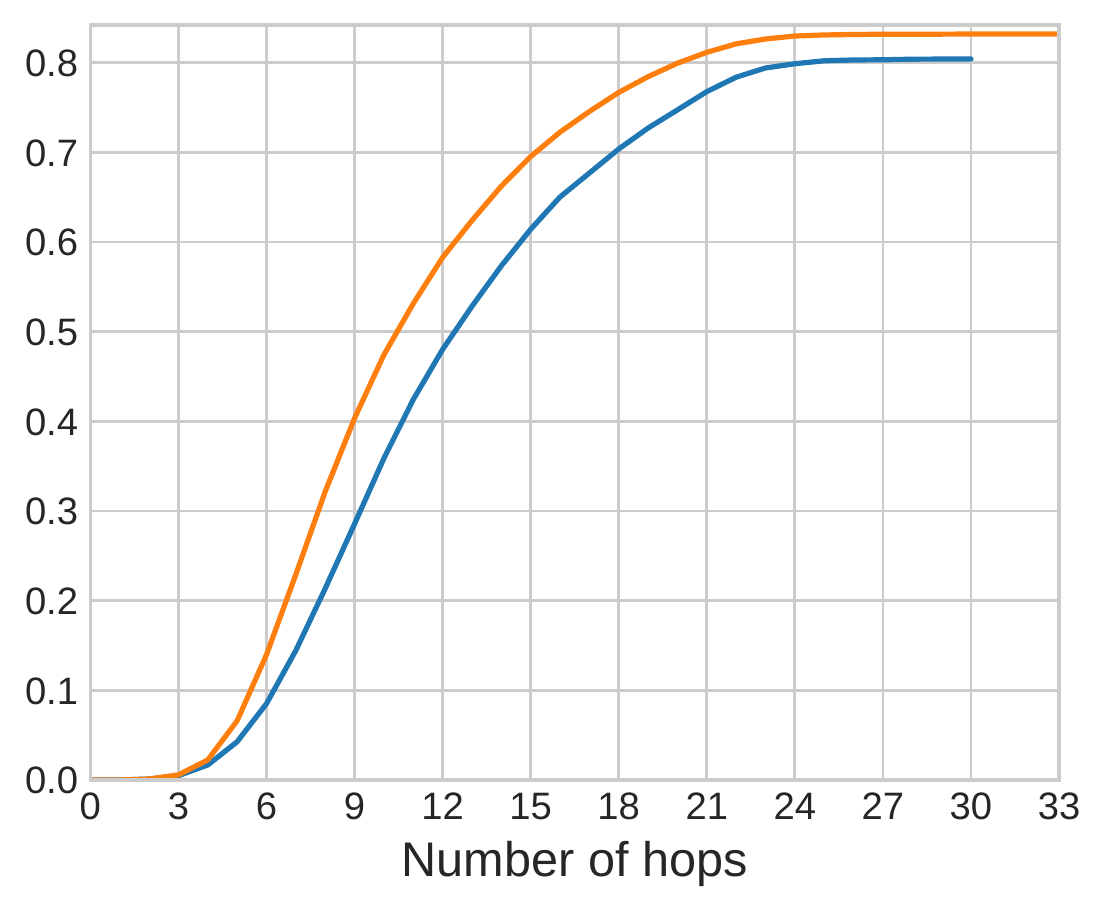}&
\includegraphics[height=4.4cm]{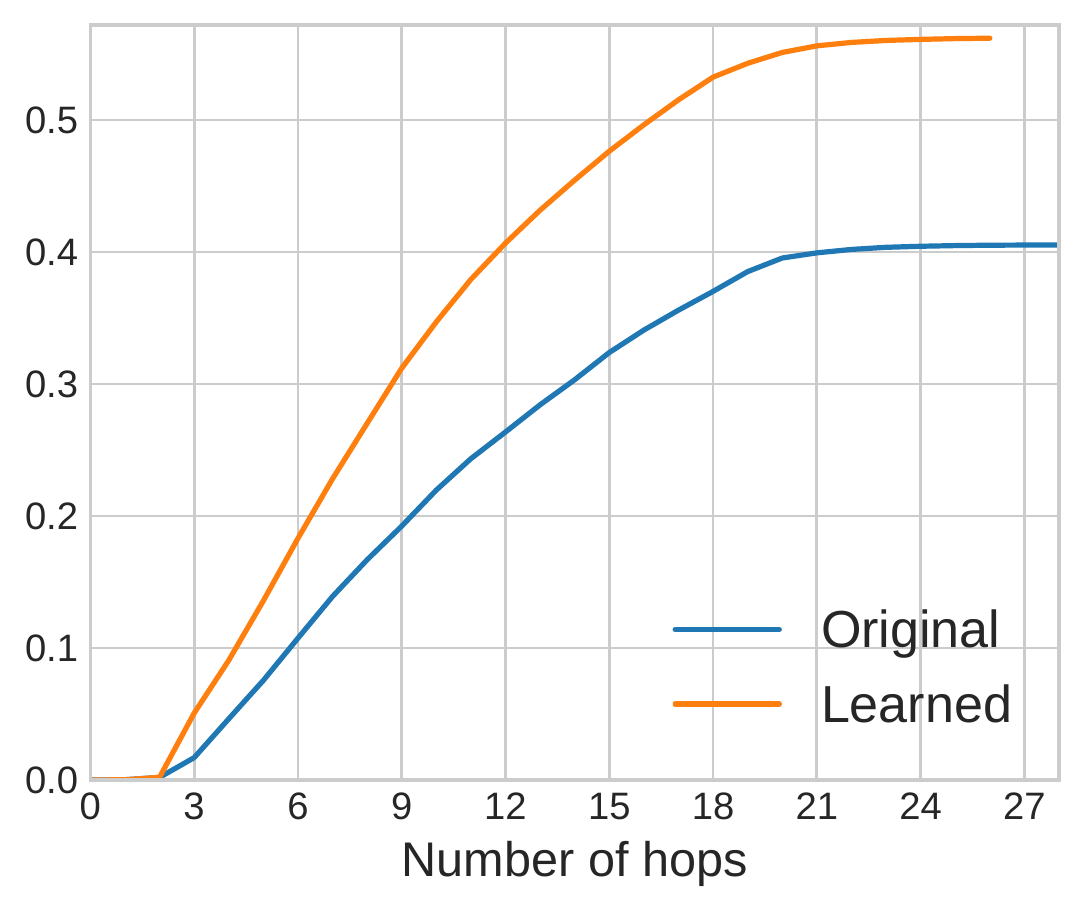}
\end{tabular}
\vspace{-7mm}
\caption{The rates of queries, for which the actual nearest neighbor was successfully found, as a function of hops made by the search algorithm for $DCS{=}128$ and $DCS{=}256$. On all the datasets the learned representations provide much higher routing quality.}
\label{fig:routing}
\vspace{-4mm}
\end{figure*}

In the first series of experiments we quantify the routing improvement from using the learned representations instead of the original datapoints. Here we consider $128$ and $256$ distance computation budgets and do not perform dimensionality reduction, $g_\theta(q) = q$. In \fig{routing} we provide the percentage of queries, for which the actual nearest neighbor was successfully found, as a function of hops made by the search algorithm. For all the datasets, the learned representations provide much better routing, especially for an extreme budget of $128$ distance computations. E.g. on SIFT100K the search algorithm reaches about $15\%$ and $12\%$ higher successful search rate for $DCS=128$ and $DCS=256$ respectively.

\subsection{Search performance}
As the most important experiment, we compare the performance of NSW graphs, using the routing based on the original datapoints and the learned representations. In more details, we perform the following comparisons:
\vspace{-3mm}

\begin{itemize}[leftmargin=2mm,itemindent=.5cm,labelwidth=\itemindent,labelsep=0cm,align=left]
\itemsep-0.15em
\item In the case without dimensionality reduction $g_\theta(q) = q$ we compare our method with the NSW graph that uses the routing on the original datapoints.

\item In the case with dimensionality reduction $g_\theta(q) = W\times q, W \in \mathbf{R}^{d\times D}$ our approach is compared to the following baseline. We compress the base vectors by PCA (trained on base set) to dimensionality $d$, then construct a new NSW graph on the truncated vectors. During the search we make the routing decisions based on the truncated vectors and rerank top-K results based on the $D$-dimensional vectors.
\end{itemize}
\vspace{-2mm}

We perform the routing on the vector representations using the NSW graph and collect a candidate list of length $k$ until the budget of distance computations is exceeded. Then the candidates are reranked by distances between the query and the original vectors. Finally, the best candidate is returned as an answer. Note that we do not need to perform reranking when routing on the original vectors. Thus, all the compared methods reserve $k$ distance computations for the reranking stage except for the NSW on the original datapoints. 

In the scenario with dimensionality reduction, we experiment with $\times2$ and $\times4$ compression rates $C$. In this case the methods additionally reserve $d$ distance computations for the matrix-vector multiplication. Thus, the budget of distance computations solely for routing equals $rDCS=(DCS - k)$ for the fully-dimensional case and $rDCS=C\times(DCS - k - d)$ for the case with the dimensionality reduction. All the results are presented in \tab{main} along with the corresponding values of $k$ and $rDCS$. Below we highlight some key observations:

\vspace{-3mm}
\begin{itemize}[leftmargin=2mm,itemindent=.5cm,labelwidth=\itemindent,labelsep=0cm,align=left]
\itemsep-0.1em
    \item First, we compare the routing performance without dimensionality reduction. For all three datasets the search on the learned representations with $k{=}8$ substantially outperforms the routing on the original datapoints. For instance, for $DCS=128$ the proposed approach reaches up to $13$ percent points improvement in $R@1$.
    \item The routing on the compressed representations, followed by reranking, demonstrates significantly higher recall rates than the routing on full-size representations if $d$ is sufficiently smaller than the $DCS$ budget, i.e. $rDCS$ does not become too small. In most of the operating points we observe the improvement from the usage of compressed representations for routing (both PCA and ours) within the same $DCS$ budget. This implies that the compressed vectors preserve the information to collect precise candidate lists, while allowing to visit more vertices within the same computational cost. 
    \item The benefits from the proposed learnable routing are more impressive in aggressive operating points of low computational budgets. However, as we show on \fig{routing}, for all the budgets the learnable routing reaches a groundtruth vertex faster.
    \item The learned low-dimensional representations reveal better routing quality compared to the PCA-truncated vectors. For $DCS=128$ and $DCS=256$ the usage of the learned representations leads to substantial increase of the search performance especially on DEEP100K and GloVe100K. E.g. on GloVe100K the performance on our low-dimensional vectors is up to $21\%$ higher compared to the routing on the PCA-truncated datapoints.

    

\end{itemize}

\begin{table*}
\small
\centering
\addtolength{\tabcolsep}{2pt}
\renewcommand\arraystretch{1.0}

\begin{tabular}{|c|c|ccc|ccc|ccc|}

\hline
Total DCS & Vertex & \multicolumn{3}{|c|}{SIFT100K} & 
\multicolumn{3}{|c|}{DEEP100K} &
\multicolumn{3}{|c|}{GloVe100K}\\\cline{3-11}
budget & representations &  
{\fontsize{8}{0}\selectfont k} & 
{\fontsize{8}{0}\selectfont rDCS} & 
{\fontsize{8}{0}\selectfont R@1} & 
{\fontsize{8}{0}\selectfont k} & 
{\fontsize{8}{0}\selectfont rDCS} & 
{\fontsize{8}{0}\selectfont R@1} &   
{\fontsize{8}{0}\selectfont k} & 
{\fontsize{8}{0}\selectfont rDCS} & 
{\fontsize{8}{0}\selectfont R@1}
\\
\hline
& Original & 0 & 128 & 0.239 & 0 & 128 & 0.386	
& 0 & 128 & 0.198 \\
& Ours & 8 & 120 & 0.371 & 8 & 120 & 0.474	
& 8 & 120 & 0.305 \\
128 & PCA$\times2$  & 8 & 112 & 0.180 & 8 & 144 & 0.399	
& - & - & - \\
& Ours$\times2$ & 8 & 112 & 0.311 & 8 & 144 & 0.565	
& - & - & - \\
& PCA$\times4$ & 16 & 320 & 0.794 & 32 & 288 & 0.673	
& 8 & 180 & 0.150 \\
& Ours$\times4$ & 16 & 320 & \textbf{0.837} & 32 & 288 & \textbf{0.779}	
& 8 & 180 & \textbf{0.343} \\
\hline
& Original & 0 & 256 & 0.672 & 0 & 256 & 0.795	
& 0 & 256 & 0.400 \\
& Ours & 8 & 248 & 0.799 & 8 & 248 & 0.811	
& 8 & 248 & 0.526 \\
256 & PCA$\times2$ & 16 & 352 & 0.855 & 16 & 384 & 0.869	
& 8 & 196 & 0.243 \\
& Ours$\times2$ & 16 & 352 & 0.893 & 16 & 384 & 0.888	
& 8 & 196 & 0.415 \\
& PCA$\times4$ & 32 & 768 & \textbf{0.965} & 64 & 672 & 0.871
& 32 & 596 & 0.394 \\
& Ours$\times4$ & 32 & 768 & 0.960 & 64 & 672 & \textbf{0.917}
& 32 & 596 & \textbf{0.604}\\
\hline
& Original & 0 & 512 & 0.936 & 0 & 512 & 0.940	
& 0 & 512 & 0.582 \\
512 & Ours & 16 & 496 & 0.949 & 16 & 496 & 0.945	
& 16 & 496 & 0.676 \\
& PCA$\times2$ & 64 & 768 & 0.980 & 64 & 800 & 0.967	
& 32 & 660 & 0.616 \\
& Ours$\times2$ & 64 & 768 & \textbf{0.981} & 64 & 800 & \textbf{0.973}	
 & 32 & 660 & \textbf{0.699} \\
\hline
\end{tabular}
\caption{The search performance $Recall@1$ for the routing based on different vertex representations. Top-$k$ candidates are collected based on the routing representations and then reranked based on the distances from the original datapoints to a query. The nearest candidate after reranking is returned as a final search answer. $rDCS$ denotes the number of distance computations the algorithm is allowed to make solely for routing purposes. $k$ and $rDCS$ values are set such that the search process performs exactly $DCS$ distance computations.}
\vspace{-3mm}
\label{tab:main}
\end{table*}

\subsection{Ablation}

In this section we compare the architectures of $f_\theta(\cdot)$ and training objectives for the proposed method. All ablation experiments were done on the GloVe100K dataset in the operating point of $DCS{=}256, k=32$ with compression to $d{=}\frac{D}{4}$. The following schemes are compared:
\begin{itemize}[leftmargin=2mm,itemindent=.5cm,labelwidth=\itemindent,labelsep=0cm,align=left]
\itemsep-0.1em
    \item \textbf{PCA}. The routing is performed on the PCA-truncated vectors. The details are discussed in section \sect{experiments}.
    \item \textbf{Ours}. Our main algorithm as described in \sect{theory} and evaluated in \sect{experiments}. For $f_\theta(\cdot)$ we use the architecture depicted on \fig{arch}. It consists of three convolutional blocks with 256 filters followed by a feed-forward network. The feed-forward network consists of two fully-connected layers with 4096 hidden units and ELU nonlinearity.
    \item \textbf{Ours + Feed-forward}. Like Ours, but $f_\theta(\cdot)$ is a feed-forward network without convolutional blocks in front.
    \item\textbf{Ours + Teacher Forcing}. Like Ours but the substitute training objective with \eq{mle}. The agent is trained on optimal routings instead of its own trajectories. 
    \item\textbf{Ours + TopK only}. Like Ours, but training objective only consists of $log P(v^* \in TopK | q, V, \theta) $ term. Hence the agent is not trained to follow the optimal routing, but only to select the best vertices.
\end{itemize}

In \sect{theory} we discuss the problem of the naive objective \eq{mle} and come to the objective \eq{imitation_objective} dictated by the Imitation Learning paradigm. In this experiment, we also provide a comparison between the optimization of these objectives. The results are presented in \tab{ablation}. When trained with Teacher Forcing objective, model achieves better objective function value but provides significantly lower recall. This result is expected since the model was not trained to cope with its errors. Surprisingly enough, training with only the TopK objective still provides competitive results. 

\begin{table}
 \centering
 \renewcommand\arraystretch{1.0}
\begin{tabular}{|c|c|}
     \hline
     Method & $Recall@1$ \\
     \hline
     PCA & 0.394\\
     \hline
     Ours & \textbf{0.604}\\
     \hline
     Ours + Feed-forward & 0.549\\
     \hline
     Ours + Teacher Forcing & 0.377 \\
     \hline
     Ours + TopK only & 0.512\\
     \hline 
\end{tabular}
\vspace{-1mm}
\caption{Ablation study for different $f_\theta(\cdot)$ and training objectives on the GloVe100K dataset. The operating point is $DCS{=}256, k{=}32$ and $\times4$ compression rate.}
\label{tab:ablation}
\vspace{-2mm}
\end{table}
\subsection{Comparison to the existing NNS methods}
Finally, we provide the comparison of our approach to the existing NNS methods for the budget $DCS{=}512$. Namely, we include in the comparison the randomized partition tree ensembles from the Annoy library\cite{annoy}, which is shown to be one of the most efficient non-graph-based method\footnote[2]{https://github.com/erikbern/ann-benchmarks}. We also report the numbers for the recent NSG graph\cite{NSG} and the multi-layer HNSW graph. The results are collected in \tab{sota}. On GloVe100K NSW on the learned representations outperforms all graphs on the original data up to $9.4\%$. Note that advantage of our method is larger for the problems of higher dimensionality. This implies that our learnable routing is more beneficial in high-dimensional spaces, where the routing problem is more challenging. On SIFT100K the highest search accuracy is achieved by the NSG graph. Note, however, that our learnable routing could also be applied to NSG and increase its performance as well.


\begin{table}
\renewcommand\arraystretch{1.0}
\begin{tabular}{|c|c|c|c|}
     \hline
     Method & SIFT100K & DEEP100K & GloVe100K\\
    \hline
     NSW+Ours & 0.949 & \textbf{0.949} & \textbf{0.676}\\
     \hline
     NSW & 0.936 & 0.940 & 0.582\\
     \hline
     NSG & \textbf{0.954} & 0.946 &  0.569\\
     \hline
     HNSW & 0.951 & 0.940 & 0.573\\
     \hline
      Annoy & 0.817 & 0.820 & 0.368\\
     \hline
\end{tabular}
\vspace{-3mm}
\caption{The experimental comparison to the existing NNS methods. We provide the $Recall@1$ values for the $DCS{=}512$ budget without dimensionality reduction.}
\label{tab:sota}
\vspace{-6mm}
\end{table}


\section{Conclusion}
\label{sect:conclusion}
In this paper we have introduced the learnable routing algorithm for NNS in similarity graphs. We propose to perform routing based on the learned vertex representations that are optimized to provide the optimal routes from the start vertex to the actual nearest neighbors. In our evaluation, we have shown that our algorithm is less susceptible to the local minima and achieves much higher recall rates under the same computational budget. The advantages of our approach come at a price of DNN training on a large set of training queries, which is performed offline and does not impose additional online costs.

\bibliography{example_paper}

\begin{thebibliography}{38}
\providecommand{\natexlab}[1]{#1}
\providecommand{\url}[1]{\texttt{#1}}
\expandafter\ifx\csname urlstyle\endcsname\relax
  \providecommand{\doi}[1]{doi: #1}\else
  \providecommand{\doi}{doi: \begingroup \urlstyle{rm}\Url}\fi

\bibitem[Andoni \& Indyk(2008)Andoni and Indyk]{andoni2008near}
Andoni, A. and Indyk, P.
\newblock Near-optimal hashing algorithms for near neighbor problem in high
  dimension.
\newblock \emph{Communications of the ACM}, 51\penalty0 (1):\penalty0 117--122,
  2008.

\bibitem[Andoni et~al.(2015)Andoni, Indyk, Laarhoven, Razenshteyn, and
  Schmidt]{Razenshteyn15}
Andoni, A., Indyk, P., Laarhoven, T., Razenshteyn, I.~P., and Schmidt, L.
\newblock Practical and optimal {LSH} for angular distance.
\newblock In \emph{NIPS}, 2015.

\bibitem[Attia \& Dayan(2018)Attia and Dayan]{imit}
Attia, A. and Dayan, S.
\newblock Global overview of imitation learning.
\newblock \emph{CoRR}, abs/1801.06503, 2018.

\bibitem[Ba et~al.(2016)Ba, Kiros, and Hinton]{ln}
Ba, J.~L., Kiros, J.~R., and Hinton, G.~E.
\newblock Layer normalization.
\newblock \emph{arXiv preprint arXiv:1607.06450}, 2016.

\bibitem[Babenko \& Lempitsky(2016)Babenko and Lempitsky]{BabenkoCVPR16}
Babenko, A. and Lempitsky, V.~S.
\newblock Efficient indexing of billion-scale datasets of deep descriptors.
\newblock In \emph{CVPR}, 2016.

\bibitem[Beaumont et~al.(2007)Beaumont, Kermarrec, and Riviere]{Beaumont07}
Beaumont, O., Kermarrec, A., and Riviere, E.
\newblock Peer to peer multidimensional overlays: Approximating complex
  structures.
\newblock In \emph{Principles of Distributed Systems, 11th International
  Conference, {OPODIS} 2007, Guadeloupe, French West Indies, December 17-20,
  2007. Proceedings}, pp.\  315--328, 2007.

\bibitem[Bentley(1975)]{KdTree}
Bentley, J.~L.
\newblock Multidimensional binary search trees used for associative searching.
\newblock \emph{Commun. ACM}, 18, 1975.

\bibitem[Bernhardsson(2012)]{annoy}
Bernhardsson, E.
\newblock Annoy: Approximate nearest neighbors in c++/python.
\newblock \url{https://github.com/spotify/annoy}, 2012.

\bibitem[Chang et~al.(2015)Chang, Krishnamurthy, Agarwal, Daum{\'e}, and
  Langford]{Daume_better}
Chang, K.-W., Krishnamurthy, A., Agarwal, A., Daum{\'e}, H., and Langford, J.
\newblock Learning to search better than your teacher.
\newblock In \emph{ICML}, 2015.

\bibitem[Chen et~al.(2018)Chen, Ma, and Xiao]{fastgcn}
Chen, J., Ma, T., and Xiao, C.
\newblock Fastgcn: Fast learning with graph convolutional networks via
  importance sampling.
\newblock \emph{CoRR}, abs/1801.10247, 2018.

\bibitem[Cheng et~al.(2017)Cheng, He, and Zhang]{scene_labelling}
Cheng, F., He, X., and Zhang, H.~J.
\newblock Stacked learning to search for scene labeling.
\newblock \emph{IEEE Transactions on Image Processing}, 26:\penalty0
  1887--1898, 2017.

\bibitem[Clevert et~al.(2015)Clevert, Unterthiner, and Hochreiter]{elu}
Clevert, D.-A., Unterthiner, T., and Hochreiter, S.
\newblock Fast and accurate deep network learning by exponential linear units
  (elus).
\newblock \emph{arXiv preprint arXiv:1511.07289}, 2015.

\bibitem[Dasgupta \& Freund(2008)Dasgupta and Freund]{RpTree}
Dasgupta, S. and Freund, Y.
\newblock Random projection trees and low dimensional manifolds.
\newblock In \emph{Proceedings of the 40th Annual {ACM} Symposium on Theory of
  Computing, Victoria, British Columbia, Canada, May 17-20, 2008}, pp.\
  537--546, 2008.

\bibitem[Dasgupta \& Sinha(2013)Dasgupta and Sinha]{dasgupta2013randomized}
Dasgupta, S. and Sinha, K.
\newblock Randomized partition trees for exact nearest neighbor search.
\newblock In \emph{Conference on Learning Theory}, pp.\  317--337, 2013.

\bibitem[Datar et~al.(2004)Datar, Immorlica, Indyk, and Mirrokni]{LSH}
Datar, M., Immorlica, N., Indyk, P., and Mirrokni, V.~S.
\newblock Locality-sensitive hashing scheme based on p-stable distributions.
\newblock In \emph{Proceedings of the 20th {ACM} Symposium on Computational
  Geometry, Brooklyn, New York, USA, June 8-11, 2004}, pp.\  253--262, 2004.

\bibitem[Daum{\'e} et~al.(2009)Daum{\'e}, Langford, and Marcu]{Daume}
Daum{\'e}, Iii, H., Langford, J., and Marcu, D.
\newblock Search-based structured prediction.
\newblock \emph{Mach. Learn.}, 75\penalty0 (3), June 2009.

\bibitem[Fu \& Cai(2016)Fu and Cai]{fu2016efanna}
Fu, C. and Cai, D.
\newblock Efanna: An extremely fast approximate nearest neighbor search
  algorithm based on knn graph.
\newblock \emph{arXiv preprint arXiv:1609.07228}, 2016.

\bibitem[Fu et~al.(2017)Fu, Xiang, Wang, and Cai]{NSG}
Fu, C., Xiang, C., Wang, C., and Cai, D.
\newblock Fast approximate nearest neighbor search with the navigating
  spreading-out graph.
\newblock \emph{arXiv preprint arXiv:1707.00143}, 2017.

\bibitem[He et~al.(2016)He, Zhang, Ren, and Sun]{he2016deep}
He, K., Zhang, X., Ren, S., and Sun, J.
\newblock Deep residual learning for image recognition.
\newblock In \emph{Proceedings of the IEEE conference on computer vision and
  pattern recognition}, pp.\  770--778, 2016.

\bibitem[Ho \& Ermon(2016)Ho and Ermon]{gail}
Ho, J. and Ermon, S.
\newblock Generative adversarial imitation learning.
\newblock In \emph{Advances in Neural Information Processing Systems 29: Annual
  Conference on Neural Information Processing Systems 2016, December 5-10,
  2016, Barcelona, Spain}, pp.\  4565--4573, 2016.

\bibitem[Indyk \& Motwani(1998)Indyk and Motwani]{LSH98}
Indyk, P. and Motwani, R.
\newblock Approximate nearest neighbors: Towards removing the curse of
  dimensionality.
\newblock In \emph{Proceedings of the Thirtieth Annual {ACM} Symposium on the
  Theory of Computing, Dallas, Texas, USA, May 23-26, 1998}, pp.\  604--613,
  1998.

\bibitem[J{\'e}gou et~al.(2011)J{\'e}gou, Douze, and Schmid]{Jegou11a}
J{\'e}gou, H., Douze, M., and Schmid, C.
\newblock Product quantization for nearest neighbor search.
\newblock \emph{TPAMI}, 33\penalty0 (1), 2011.

\bibitem[Kingma \& Ba(2014)Kingma and Ba]{kingma2014adam}
Kingma, D.~P. and Ba, J.
\newblock Adam: A method for stochastic optimization.
\newblock \emph{arXiv preprint arXiv:1412.6980}, 2014.

\bibitem[Kipf \& Welling(2016)Kipf and Welling]{gcn}
Kipf, T.~N. and Welling, M.
\newblock Semi-supervised classification with graph convolutional networks.
\newblock \emph{arXiv preprint arXiv:1609.02907}, 2016.

\bibitem[Malkov \& Yashunin(2016)Malkov and Yashunin]{malkov2018efficient}
Malkov, Y.~A. and Yashunin, D.
\newblock Efficient and robust approximate nearest neighbor search using
  hierarchical navigable small world graphs.
\newblock \emph{arXiv preprint arXiv:1603.09320}, 2016.

\bibitem[McCartin{-}Lim et~al.(2012)McCartin{-}Lim, McGregor, and
  Wang]{ApdTree}
McCartin{-}Lim, M., McGregor, A., and Wang, R.
\newblock Approximate principal direction trees.
\newblock In \emph{Proceedings of the 29th International Conference on Machine
  Learning, {ICML} 2012, Edinburgh, Scotland, UK, June 26 - July 1, 2012},
  2012.

\bibitem[Navarro(2002)]{navarro2002searching}
Navarro, G.
\newblock Searching in metric spaces by spatial approximation.
\newblock \emph{The VLDB Journal}, 11\penalty0 (1):\penalty0 28--46, 2002.

\bibitem[Negrinho et~al.(2018)Negrinho, Gormley, and Gordon]{bso_2018_bad}
Negrinho, R., Gormley, M.~R., and Gordon, G.~J.
\newblock Learning beam search policies via imitation learning.
\newblock In \emph{Proceedings of {NIPS}}, 2018.

\bibitem[Pennington et~al.(2014)Pennington, Socher, and
  Manning]{pennington2014glove}
Pennington, J., Socher, R., and Manning, C.~D.
\newblock Glove: Global vectors for word representation.
\newblock In \emph{Empirical Methods in Natural Language Processing (EMNLP)},
  2014.

\bibitem[Ross et~al.(2011)Ross, Gordon, and Bagnell]{dagger}
Ross, S., Gordon, G.~J., and Bagnell, D.
\newblock A reduction of imitation learning and structured prediction to
  no-regret online learning.
\newblock In \emph{Proceedings of the Fourteenth International Conference on
  Artificial Intelligence and Statistics, {AISTATS} 2011, Fort Lauderdale, USA,
  April 11-13, 2011}, pp.\  627--635, 2011.

\bibitem[Shapiro(1987)]{Bisiani87}
Shapiro, S.~C.
\newblock \emph{Encyclopedia of Artificial Intelligence}.
\newblock 1987.

\bibitem[Smith \& Topin(2017)Smith and Topin]{smith2017super}
Smith, L.~N. and Topin, N.
\newblock Super-convergence: Very fast training of residual networks using
  large learning rates.
\newblock \emph{arXiv preprint arXiv:1708.07120}, 2017.

\bibitem[Sproull(1991)]{PcaTree}
Sproull, R.~F.
\newblock Refinements to nearest-neighbor searching in k-dimensional trees.
\newblock \emph{Algorithmica}, 6, 1991.

\bibitem[Williams \& Zipser(1989)Williams and Zipser]{williams1989learning}
Williams, R.~J. and Zipser, D.
\newblock A learning algorithm for continually running fully recurrent neural
  networks.
\newblock \emph{Neural computation}, 1\penalty0 (2):\penalty0 270--280, 1989.

\bibitem[Wiseman \& Rush(2016{\natexlab{a}})Wiseman and Rush]{bso_good}
Wiseman, S. and Rush, A.~M.
\newblock Sequence-to-sequence learning as beam-search optimization.
\newblock In \emph{EMNLP}, 2016{\natexlab{a}}.

\bibitem[Wiseman \& Rush(2016{\natexlab{b}})Wiseman and
  Rush]{wiseman2016sequence}
Wiseman, S. and Rush, A.~M.
\newblock Sequence-to-sequence learning as beam-search optimization.
\newblock \emph{arXiv preprint arXiv:1606.02960}, 2016{\natexlab{b}}.

\bibitem[Yu et~al.(2017)Yu, Hsieh, Lei, and Dhillon]{yu2017greedy}
Yu, H.-F., Hsieh, C.-J., Lei, Q., and Dhillon, I.~S.
\newblock A greedy approach for budgeted maximum inner product search.
\newblock In \emph{Advances in Neural Information Processing Systems}, pp.\
  5453--5462, 2017.

\bibitem[Zhou et~al.(2018)Zhou, Cui, Zhang, Yang, Liu, and Sun]{GNN}
Zhou, J., Cui, G., Zhang, Z., Yang, C., Liu, Z., and Sun, M.
\newblock Graph neural networks: {A} review of methods and applications.
\newblock \emph{CoRR}, abs/1812.08434, 2018.

\end{thebibliography}
\bibliographystyle{icml2019}

\end{document}